\documentclass[letterpaper, 10 pt, conference]{ieeeconf}  % Comment this line out if you need a4paper

\usepackage{amsmath,amsfonts}
\usepackage{algorithmic}
\usepackage{array}
\usepackage[caption=false,font=normalsize,labelfont=sf,textfont=sf]{subfig}
\usepackage{textcomp}
\usepackage{stfloats}
\usepackage{url}
\usepackage{verbatim}
\usepackage{graphicx}
\usepackage{booktabs}
\usepackage{amssymb}
\usepackage{color}
\usepackage{soul}
\sethlcolor{yellow}
\graphicspath{{figures/}}
\usepackage{rotating}
\usepackage{pifont}
\usepackage{colortbl}
\usepackage{array} % 用于扩展表格功能
\usepackage{multirow}
\usepackage{graphicx}
\usepackage[table,xcdraw]{xcolor}
\usepackage{makecell} % 调整表格

\usepackage{hyperref}

\newcommand{\cmark}{\ding{51}} % 对
\newcommand{\xmark}{\ding{55}} % 叉
\newcommand{\first}[1]{{\textbf{#1}}}

\definecolor{barriercolor}{RGB}{112, 128, 144}
\definecolor{bicyclecolor}{RGB}{220, 20, 60}
\definecolor{buscolor}{RGB}{255, 127, 80}
\definecolor{carcolor}{RGB}{255, 158, 0}
\definecolor{constructcolor}{RGB}{233, 150, 70}
\definecolor{motorcolor}{RGB}{255, 61, 99}
\definecolor{pedestriancolor}{RGB}{0, 0, 230}
\definecolor{trafficcolor}{RGB}{47, 79, 79}
\definecolor{trailercolor}{RGB}{255, 140, 0}
\definecolor{truckcolor}{RGB}{255, 99, 71}
\definecolor{drivablecolor}{RGB}{0, 207, 191}
\definecolor{otherflatcolor}{RGB}{175, 0, 75}
\definecolor{sidewalkcolor}{RGB}{75, 0, 75}
\definecolor{terraincolor}{RGB}{112, 180, 60}
\definecolor{manmadecolor}{RGB}{222, 184, 135}
\definecolor{vegetationcolor}{RGB}{0, 175, 0}
\definecolor{otherscolor}{RGB}{0, 0, 0}
\usepackage{makecell}
\usepackage{arydshln}
\usepackage{cleveref}
\usepackage{hhline}
\usepackage{float}
\IEEEoverridecommandlockouts                          % This command is only needed if 
\title{\LARGE \bf
Collaborative Learning of Local 3D Occupancy Prediction and Versatile Global Occupancy Mapping
}

\author{Shanshuai Yuan$^{1,2}$,
        Julong Wei$^{1}$,
        Muer Tie$^{1}$,
        Xiangyun Ren$^{2}$,
        Zhongxue Gan$^{1}$,
        and Wenchao Ding$^{1*}$
\thanks{This work was supported in part by the National Natural Science Foundation of China (NSFC) under Grant 62403142, in part by the Science and Technology Commission of Shanghai Municipality under Grant 24511103100, and in part by the State Key Laboratory of Intelligent Vehicle Safety Technology under Grant IVSTSKL-202317. The computations in this research were performed using the CFFF platform of Fudan University.}% <-this % stops a space
\thanks{Project page: \url{https://ss-yuan.github.io/LMPOcc/}.}
\thanks{$^{1}$College of Intelligent Robotics and Advanced Manufacturing, Fudan University, Shanghai, China. {\tt\small \{ssyuan23\}@m.fudan.edu.cn}, {\tt\small \{ganzhongxue,
dingwenchao\}@fudan.edu.cn}}%
\thanks{$^{2}$State Key Laboratory of Intelligent Vehicle Safety Technology, Chongqing Changan Automobile Co., Ltd., Chongqing, China.}%
\thanks{*Corresponding authors: Wenchao Ding}%
}

\begin{document}

\maketitle
\thispagestyle{empty}
\pagestyle{empty}

%%%%%%%%%%%%%%%%%%%%%%%%%%%%%%%%%%%%%%%%%%%%%%%%%%%%%%%%%%%%%%%%%%%%%%%%%%%%%%%%
\begin{abstract}

Vision-based 3D semantic occupancy prediction is vital for autonomous driving, enabling unified modeling of static infrastructure and dynamic agents. Global occupancy maps serve as long-term memory priors, providing valuable historical context that enhances local perception. This is particularly important in challenging scenarios such as occlusion or poor illumination, where current and nearby observations may be unreliable or incomplete.
Priors aggregated from previous traversals under better conditions help fill gaps and enhance the robustness of local 3D occupancy prediction.
In this paper, we propose Long-term Memory Prior Occupancy (LMPOcc), a plug-and-play framework that incorporates global occupancy priors to boost local prediction and simultaneously updates global maps with new observations. 
To realize the information gain from global priors, we design an efficient and lightweight Current-Prior Fusion module that adaptively integrates prior and current features. Meanwhile, we introduce a model-agnostic prior format to enable continual updating of global occupancy and ensure compatibility across diverse prediction baselines.
LMPOcc achieves state-of-the-art local occupancy prediction performance validated on the Occ3D-nuScenes benchmark, especially on static semantic categories. 
Furthermore, we verify LMPOcc’s capability to build large-scale global occupancy maps through multi-vehicle crowdsourcing, and utilize occupancy-derived dense depth to support the construction of 3D open-vocabulary maps. 
Our method opens up a new paradigm for continuous global information updating and storage, paving the way towards more comprehensive and scalable scene understanding in large outdoor environments.
% Project page: \href{https://ss-yuan.github.io/LMPOcc/}{LMPOcc}.

\end{abstract}

%%%%%%%%%%%%%%%%%%%%%%%%%%%%%%%%%%%%%%%%%%%%%%%%%%%%%%%%%%%%%%%%%%%%%%%%%%%%%%%%
\section{INTRODUCTION}

\begin{figure}[t]
    \centering
    \Huge
    \includegraphics[width=\columnwidth]{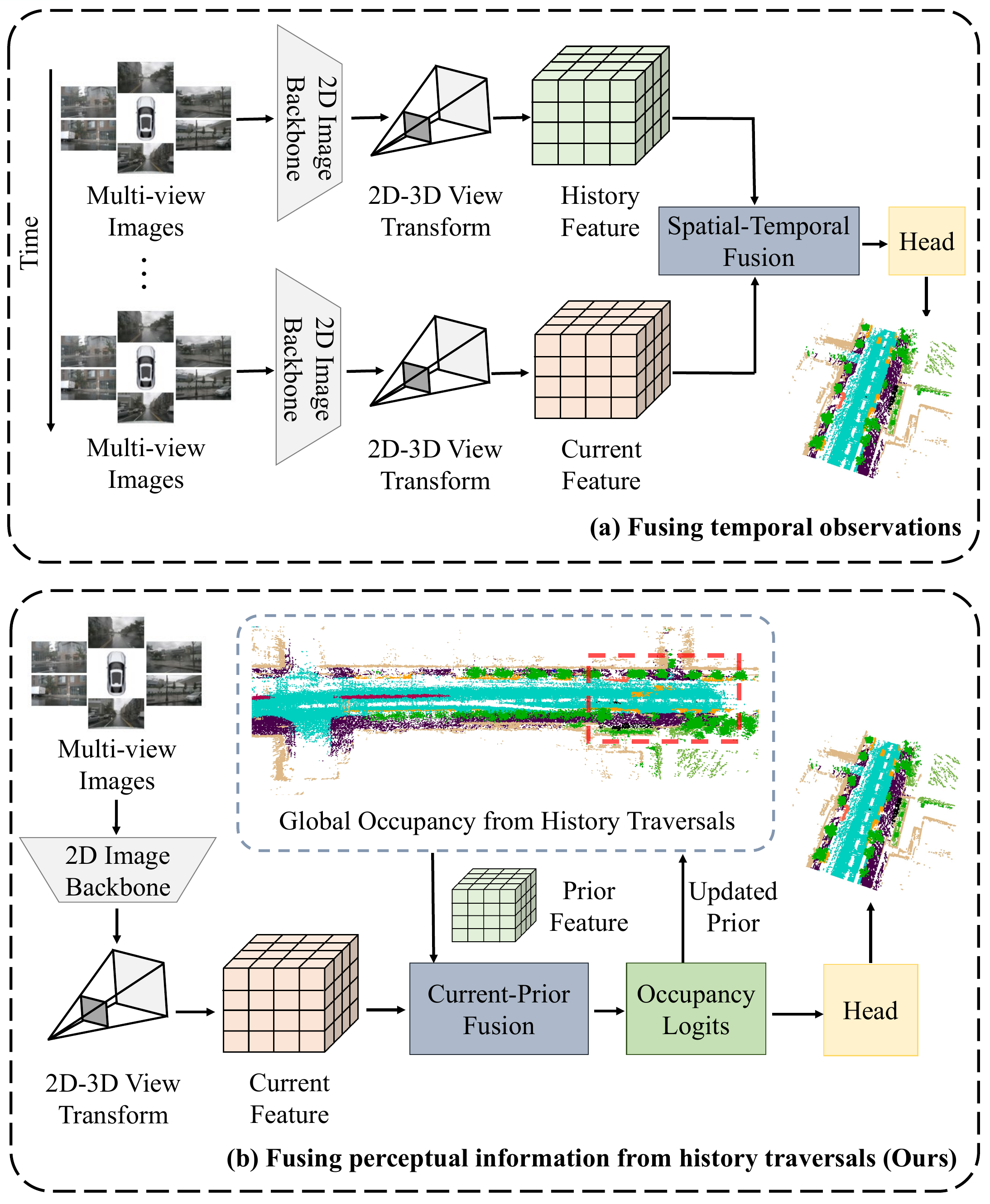}
    \vspace{-20pt}
    \caption{Comparison of temporal information integration methods in 3D occupancy prediction. (a) Existing works primarily integrate information from adjacent observations. (b) Our work fuses perceptual information obtained from historical traversals of the
    current location. The historical perceptual information constructs global occupancy and serves as long-term memory priors.}
    \label{fig:cover}
    \vspace{-9pt}
\end{figure}

%%%%%%%%%%%%%%%%%%%%%%%%%%%%%%%%%%%%%%%%%%%%
Vision-based 3D semantic occupancy prediction is fundamental for autonomous driving systems, enabling precise and unified understanding of both static infrastructure and dynamic agents. However, perception quality often varies significantly in complex real-world environments due to dynamic factors such as weather, illumination changes, and occlusions. These factors cause local sensor observations to be unreliable or incomplete, limiting robust 3D occupancy prediction.
Existing works address this issue by fusing temporal information through techniques like BEV feature alignment, self-attention mechanisms, and 3D convolution-based fusion, which primarily integrate features from adjacent frames, as illustrated in Fig.~\ref{fig:cover}(a). While effective in some situations, these methods remain vulnerable when consecutive observations share similar adverse conditions such as severe occlusion or poor lighting, leading to degraded performance.

%%%%%%%%%%%%%%%%%%%%%%%%%%%%%%%%%%%%%%%%%%
%%%%%%%%%%%%%%%%%%%%%%%%%%%%%%%%%%%%%%%%%%

To overcome these limitations, long-term memory priors play a vital role. By aggregating perceptual information collected from repeated traversals at the same geographic locations under more favorable conditions, long-term memory priors provide rich historical context that complements and corrects deficient real-time observations. These priors naturally accumulate to form a global occupancy map, which serves as persistent memory across spatial and temporal dimensions and enables accumulation and refinement of scene knowledge beyond single-instance observations, as shown in Fig.~\ref{fig:cover}(b).

Beyond enhancing local perception, the global occupancy map offers several important benefits. It provides finer and denser geometric details than raw LiDAR data, delivering high-quality dense depth information essential for many applications. A typical example is the construction of 3D open-vocabulary maps, which support flexible and scalable scene understanding. Foundation models have achieved great success mainly on 2D images. However, to effectively leverage them in large-scale outdoor 3D scenes, reliable and dense depth information is necessary. The occupancy map reflects the scene geometry and thus provides accurate depth cues. 
Since real-world environments inevitably change over time, maintaining and updating the global map through continuous local occupancy predictions is important for adapting to scene dynamics.

In this paper, we propose Long-term Memory Prior Occupancy (LMPOcc), a novel framework that jointly performs local occupancy perception and constructs a global occupancy map. 
Specifically, our framework leverages ego-to-global coordinate transformation to simultaneously construct global occupancy representations from localized perceptual outputs, and utilize corresponding spatial memory priors from the global map to refine real-time local inference through cross-temporal feature alignment. 
We store visibility-region occupancy logits from each local prediction into the global map. This model-agnostic prior format enables crowdsourced construction of city-level global occupancy. To fully leverage these long-term memory priors, we design an efficient and lightweight Current-Prior Fusion module that learns adaptive weights between current and prior features to produce refined occupancy predictions. 
We employ ray casting to extract dense depth from occupancy maps, providing high-quality depth data for 3D open-vocabulary map construction. Extensive experiments on the Occ3D-nuScenes~\cite{tian2023occ3d} benchmark demonstrate that LMPOcc significantly improves 3D occupancy prediction baselines and achieves state-of-the-art performance.

The main contributions of this work are summarized as follows:

• We propose LMPOcc, the first framework that leverages global occupancy as a long-term memory prior to enhance local 3D occupancy prediction while simultaneously constructing and updating the global map. Our method also provides dense depth information to support large-scale outdoor applications such as 3D open-vocabulary mapping.

• We design a plug-and-play architecture to realize the bidirectional interaction between global and local occupancy. In particular, we introduce a model-agnostic prior format and develop an efficient lightweight Current-Prior Fusion module for cross-temporal feature integration.

• Validated on the Occ3D-nuScenes benchmark, LMPOcc achieves state-of-the-art performance. We demonstrate its capability to build large-scale global occupancy maps via multi-vehicle crowdsourcing, and to leverage occupancy-derived dense depth for 3D open vocabulary map construction.

%%%%%%%%%%%%%%%%%%%%%%%%%%%%%%%%%%%%%%%%%%%%

\section{RELATED WORKS}

\subsection{3D Semantic Occupancy Prediction}

Vision-based 3D occupancy prediction has seen significant advancements driven by a variety of methodological innovations. 
Early supervised frameworks like MonoScene~\cite{cao2022monoscene} establish 2D-3D U-Net architectures, while BEVDet~\cite{huang2021bevdet} and BEVFormer~\cite{li2024bevformer} introduce view transformation via LSS~\cite{philion2020lift} projection and spatio-temporal transformers respectively. Recent works propose strategies to enhance both representation and computational efficiency in 3D occupancy prediction. For example, TPVFormer~\cite{huang2023tri} introduces three-view perspective encoding and SurroundOcc~\cite{wei2023surroundocc} employs multi-scale refinement for spatial detail capture. Unsupervised approaches like SelfOcc~\cite{huang2024selfocc} and OccNeRF~\cite{zhang2023occnerf} circumvent dense labeling through neural rendering. 
To enhance computational efficiency, VoxFormer~\cite{li2023voxformer}  and OctreeOcc~\cite{lu2023octreeocc}  utilize sparse voxel representations to optimize memory and speed.
% while FastOcc~\cite{hou2024fastocc}  achieves acceleration by replacing 3D convolutions with lightweight 2D alternatives.
GaussianFusionOcc~\cite{pavkovic2025gaussianfusionocc} fuses multi-sensor data using semantic 3D Gaussians and deformable attention for efficient and accurate 3D occupancy prediction.

\subsection{Memory Fusion for 3D Perception}

Current memory fusion methods for 3D perception can be divided into three paradigms, involving different ways of processing history information. Attention-based approaches~\cite{li2024bevformer} leverage transformers for implicit temporal modeling, demonstrating strong dependency capture while lacking explicit geometric constraints. Geometry-aligned fusion methods exemplified by BEVDet4D~\cite{huang2022bevdet4d} and PanoOcc~\cite{wang2024panoocc} employ spatial feature alignment through estimated camera poses coupled with concatenation-convolution operations, achieving computational efficiency at the expense of long-term temporal consistency. Emerging cost volume techniques address these limitations through geometric depth reasoning, as demonstrated by SOLOFusion~\cite{park2022time} in image-space fusion and CVT-Occ~\cite{ye2024cvt} in 3D voxel adaptation.
ST-Occ~\cite{leng2025occupancy} proposes a scene-centered spatiotemporal memory to efficiently aggregate historical occupancy features from adjacent scenes.
The methods discussed above primarily integrate information from adjacent observations, while our work leverages perceptual information acquired from historical traversals of identical geographic locations.

\section{APPROACH}

\begin{figure*}[t]
    \centering
    \includegraphics[width=1\linewidth]{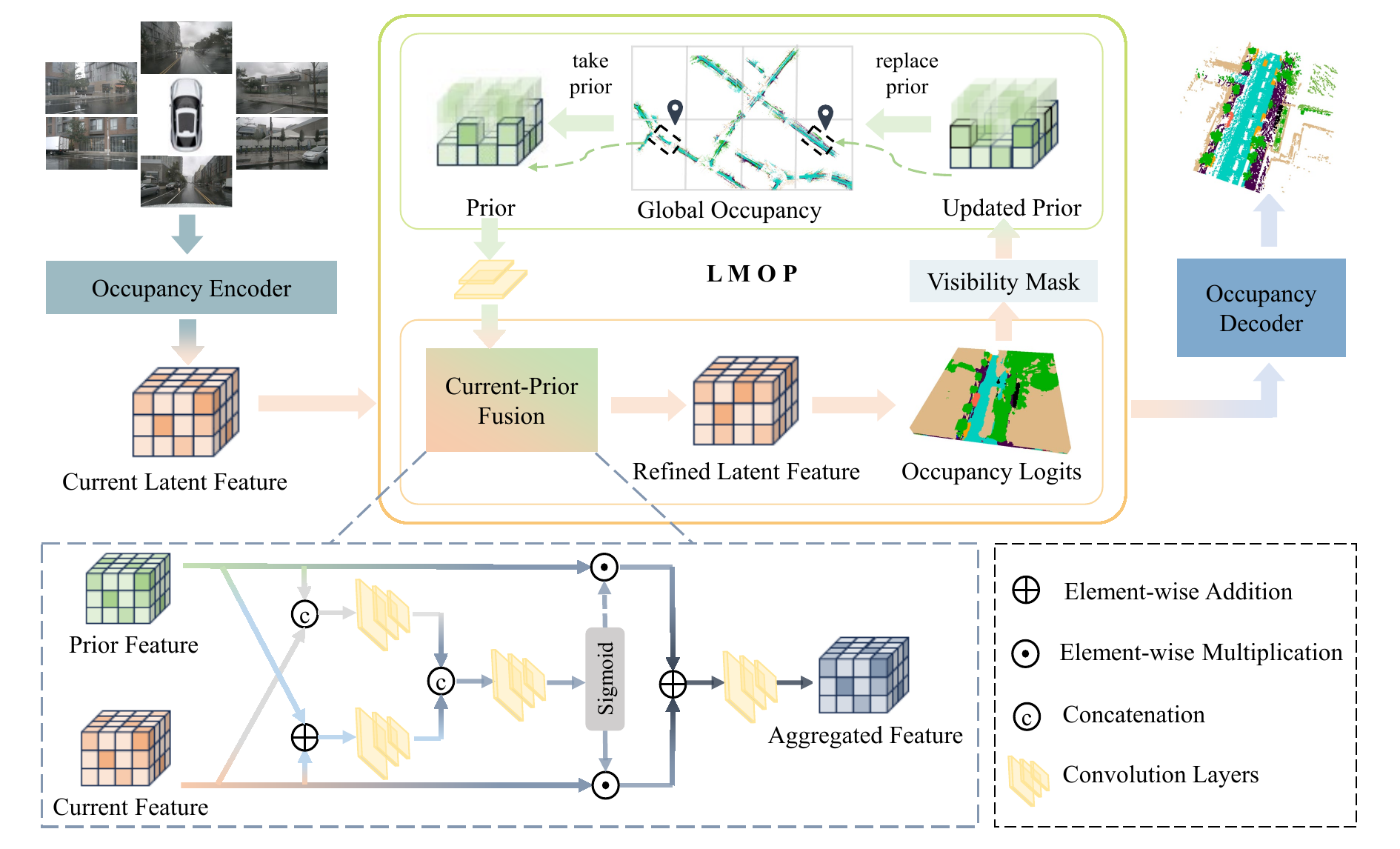}
    \vspace{-9pt}
    \caption{An overview of our LMPOcc framework. LMPOcc fristly generates Current Latent Features from surround-view images. Then it extracts spatially-aligned Prior Features from global occupancy and integrates them via the Current-Prior Fusion Module to generate Refined Latent Features. The refined latent features decode current occupancy logits, which are stored into corresponding locations in the global occupancy after visibility masking. Existing occupancy priors at these locations are replaced by the updated logits. Finally, the occupancy logits are converted into local current occupancy prediction results.}
    \label{fig:framework}
    \vspace{-9pt}
\end{figure*}

\subsection{Overall Architecture}
An overview of the Long-term Memory Prior Occupancy (LMPOcc) is presented in  Fig.~\ref{fig:framework}.
LMPOcc extends the occupancy prediction baseline by incorporating the Long-term Memory Occupancy Priors (LMOP) module (see Sec. \ref{sec:3.1}), which strengthens local perception and facilitates the construction of global occupancy.
The system receives inputs $ \mathcal{I} = \{\mathbf{I}, \mathbf{G}_{ego}\} $, comprising surround-view images $\mathbf{I}$ and the ego vehicle's local-to-global coordinate transformation $ \mathbf{G}_{ego} \in \mathbb{R}^{4 \times 4} $. The model processes surround-view images through an occupancy encoder to generate latent features. These latent features are then fed into the LMOP module to obtain enhanced occupancy logits, which are subsequently processed by an occupancy decoder to produce the final 3D semantic occupancy prediction results. In the LMOP module, the current features are fused with corresponding prior features through the Current-Prior Fusion module, yielding the refined latent features (see Sec. \ref{sec:3.2}). These refined latent features are subsequently transformed into occupancy logits via neural network processing. The occupancy logits are utilized to update prior features and generate the final occupancy prediction results. The model-agnostic prior format is introduced in Sec. \ref{sec:3.3}.
Moreover, LMPOcc also supports the construction of large-scale 3D open vocabulary maps by leveraging the dense depth information derived from occupancy results (detailed in Sec.~\ref{sec:3.4}).

\subsection{Long-Term Memory Occupancy Priors}
\label{sec:3.1}
The Long-term Memory Occupancy Priors (LMOP) module is plug-and-play and compatible with various 3D occupancy baselines. It enables local perception and global occupancy to reinforce each other.
Inspired by NMP~\cite{xiong2023neural}, the global occupancy adopts a sparse map tile structure, where each tile corresponds to a geographically aligned patch in the global coordinate system and is initialized as empty. This sparse structure significantly reduces memory usage by only storing tiles covering navigable urban zones (e.g., roads and accessible areas), thus avoiding the need to store the entire city map. Tiles can be efficiently retrieved and updated via coordinate-based indexing, allowing vehicles to load relevant local map areas on demand.
For each tile, we maintain and iteratively update the global occupancy representation $ \mathbf{P} \in \mathbb{R}^{H_{G} \times W_{G} \times (Z \cdot N_{\text{sem}})} $, where $ H_{G} $ and $ W_{G} $ define the spatial resolution of the city-level map tile, $ Z $ denotes vertical discretization depth, and $ N_{\text{sem}} $ corresponds to the number of distinguishable object categories. 
Both the global map and the local prior feature are represented in Bird's-Eye View (BEV) format through height-to-channel transformation, as expressed in FlashOcc~\cite{yu2023flashocc}. 
This BEV representation not only reduces storage costs by effectively stacking height information into channels, but also enhances bidirectional local-global indexing efficiency.
The local coordinate of each pixel in the BEV feature $ c_{t} \in \mathbb{R}^{H \times W \times 2} $ is transformed to the corresponding global coordinate $ p_{t} \in \mathbb{R}^{H \times W \times 2} $ through $ \mathbf{G}_{ego} \in \mathbb{R}^{4 \times 4} $.
This transformation ensures spatial alignment between locally perceived features and the global map tiles.
Establishing spatial correspondence between local and global occupancy, we align prior and current feature channels via convolutional layers, and then fuse current features with prior features to enhance local perception. 
The enhanced perceptual outputs, represented as occupancy logits, serve as updated priors and replace the corresponding regions within the global occupancy map. This incremental update process allows the system to refine the global occupancy over time using new local observations, thus maintaining a persistent yet dynamically evolving city-level occupancy prior.

\subsection{Current-Prior Fusion}
\label{sec:3.2}

Our Current-Prior Fusion (CPFusion) is shown in Fig.~\ref{fig:framework}. 
The CPFusion module incorporates two parallel branches, comprising a concatenation branch and an element-wise addition branch.
The concatenation branch concatenates the current feature $\boldsymbol{F}_\mathrm{c}$  and prior feature $\boldsymbol{F}_\mathrm{p}$ to form a combined feature $\boldsymbol{F}_\mathrm{cat}$, as shown in Eq.~\ref{eq:concat}. Concurrently, the element-wise addition branch gets their element-wise sum results $\boldsymbol{F}_\mathrm{add}$, as shown in Eq.~\ref{eq:add}.
These two features, $\boldsymbol{F}_\mathrm{cat}$ and $\boldsymbol{F}_\mathrm{add}$, are then concatenated and passed through a convolution layer followed by a sigmoid activation function, yielding a
tensor $\boldsymbol{\alpha}$ with values constrained between 0 and 1, as shown in Eq.~\ref{eq:alpha}. This tensor $\boldsymbol{\alpha}$ serves as a weighting factor to dynamically balance the contributions of the current and prior features through a weighted summation, as expressed in Eq.~\ref{eq:results}.
\begin{equation}
\boldsymbol{F}_\mathrm{cat}=\boldsymbol{W}_1(\boldsymbol{f}_\mathrm{cat}(\boldsymbol{F}_\mathrm{c},\boldsymbol{F}_\mathrm{p})),
\label{eq:concat}
\end{equation}
%%%%%%%%%%%%%%%%%%%%%%%%%%%%%
\begin{equation}
\boldsymbol{F}_\mathrm{add}=\boldsymbol{W}_2(\boldsymbol{F}_\mathrm{c} + \boldsymbol{F}_\mathrm{p}),
\label{eq:add}
\end{equation}
%%%%%%%%%%%%%%%%%%%%%%%%%%%
\begin{equation}
\boldsymbol{\alpha}=\sigma( 
\boldsymbol{W}_3(\boldsymbol{f}_\mathrm{cat}(\boldsymbol{F}_\mathrm{cat},\boldsymbol{F}_\mathrm{add})) ),
\label{eq:alpha}
\end{equation}
%%%%%%%%%%%%%%%%%%%%%%%%%%%%
\begin{equation}
\boldsymbol{F}_\mathrm{agg}=\boldsymbol{\alpha} \odot \boldsymbol{F}_\mathrm{c} + (1-\boldsymbol{\alpha}) \odot \boldsymbol{F}_\mathrm{p},
\label{eq:results}
\end{equation}
%%%%%%%%%%%%%%%%%%%%%%%%%%%%%%%
where $f_\mathrm{cat}(\cdot)$ and $\sigma(\cdot)$ refer to the concatenation and sigmoid, respectively. 
$\boldsymbol{W}_1$, $\boldsymbol{W}_2$ and $\boldsymbol{W}_3$ denote convolution layers. $\boldsymbol{F}_\mathrm{agg}$ represents the output features of CPFusion.

\subsection{Model-Agnostic Prior Format}
\label{sec:3.3}

The prior is stored as occupancy logits, ensuring that the global occupancy prior remains agnostic to any specific occupancy prediction model during deployment. 
After fusing current and prior features into refined latent features, the network outputs occupancy logits $ \mathbf{O}_L \in \mathbb{R}^{H_{L} \times W_{L} \times Z  \times N_{\text{sem}}} $, where $ H_{L} $, $ W_{L} $, $Z$ denote spatial dimensions and $ N_{\text{sem}} $ denotes the number of semantic classes.
To avoid storing noise outside the visible regions into the prior, we employ the camera visibility mask to retain only the content within the observable regions of the occupancy logits. The camera visibility mask is generated by casting rays from each camera origin towards voxel centers, following the approach in Occ3D-NuScenes~\cite{tian2023occ3d}. Along each ray, the first intersected occupied voxel is labeled as ``observed'', while subsequent voxels along the same ray are marked as ``unobserved''. Any voxel not intersected by these rays is automatically assigned an ``unobserved'' status. 
The masked occupancy logits are then reshaped to $ H_{L} \times W_{L} \times (Z \cdot N_{\text{sem}}) $ and used to update the corresponding regions in the global occupancy map.
Contrary to common practice, our experiments show that filtering dynamic objects from the prior fails to improve model performance, as shown in Table~\ref{tab:remove_dynamic}. This suggests historical dynamic objects may provide effective information for local perception. Therefore, our method retains dynamic components within the prior. We further propose two strategies for dynamic object removal and discuss the impact of dynamic components within the prior in Sec.~\ref{sec:4.4}.

%%%%%%%%%%%%%%%%%%%%%%%%%%%%%%%%%%%%%%%%%%
\begin{figure}[t]
    \centering
    \Huge
    \includegraphics[width=\columnwidth]{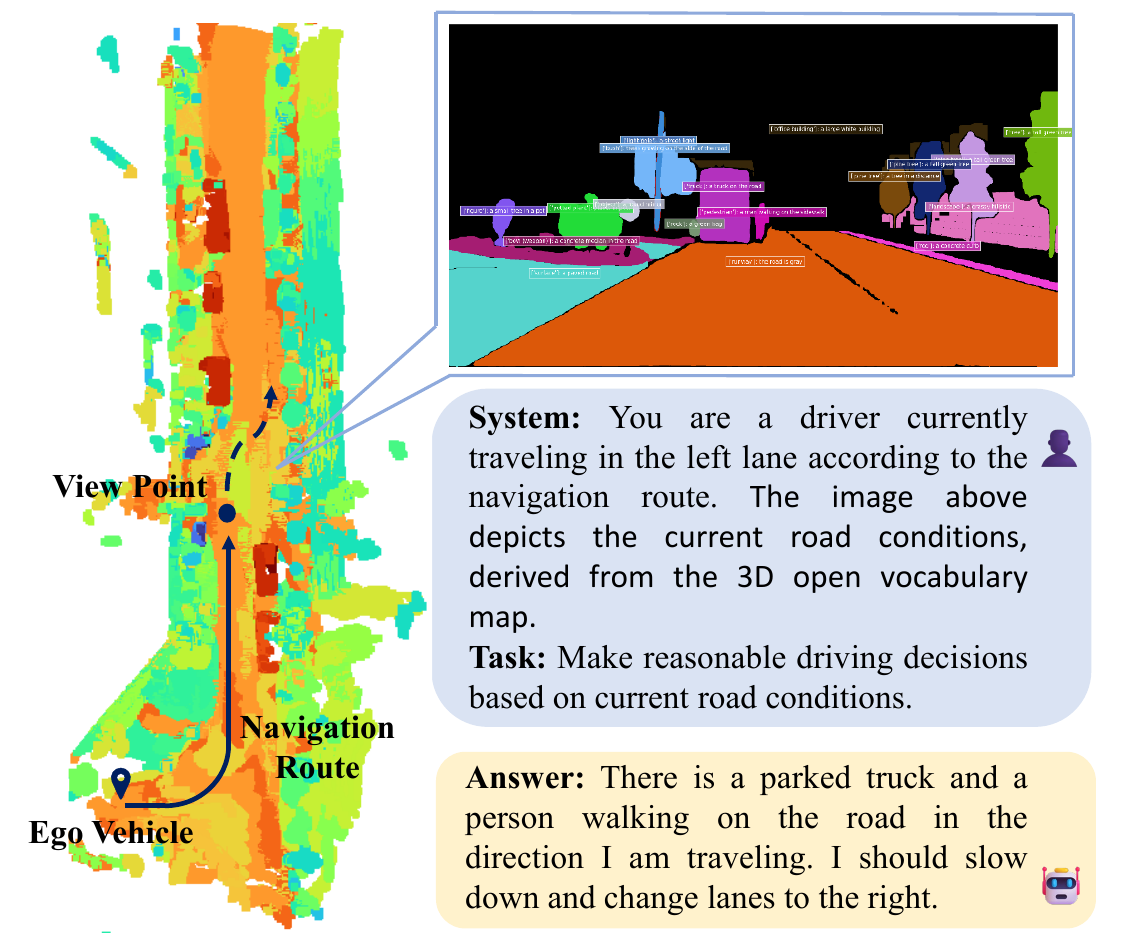}
    \vspace{-24pt}
    \caption{Demonstration of using a 3D open vocabulary map to interact with a Vision-Language Model (VLM). A viewpoint along the navigation route is selected, from which a semantic map is rendered by the 3D open vocabulary map and then interpreted by the VLM. This enables the VLM to anticipate upcoming road conditions and make informed driving decisions to handle challenging scenarios. The example dialogue shows the VLM analyzing the scene and recommending appropriate actions.}
    \label{fig:ov_map_llm}
    \vspace{-9pt}
\end{figure}
%%%%%%%%%%%%%%%%%%%%%%%%%%%%%%%%%%%%%%%%%%

%%%%%%%%%%%%%%%%%%%%%%%%%%%%%%%%%%%%%%%%%%%
\begin{table*}[h!]
\centering
\Huge
\caption{3D occupancy prediction performance on the Occ3D-nuScenes validation set. Both the small version and large version of LMPOcc outperform the models that have similar settings.}
% \vspace{-2pt}
\renewcommand\arraystretch{1.2}
\resizebox{1.0\textwidth}{!}{
\begin{tabular}{l|cc|c|c|ccccccccccccccccc}
\toprule
Method
& \begin{sideways}{History Frame}\end{sideways}
& \begin{sideways}{Resolution}\end{sideways}  
& \begin{sideways}{Backbone}\end{sideways}      
& \begin{sideways}{\textbf{mIoU} $\uparrow$}\end{sideways}  
& \begin{sideways}{\textcolor{otherscolor}{$\blacksquare$} \textbf{others}  $\uparrow$}\end{sideways}  
& \begin{sideways}{\textcolor{barriercolor}{$\blacksquare$} \textbf{barrier} $\uparrow$}\end{sideways}  
& \begin{sideways}{\textcolor{bicyclecolor}{$\blacksquare$} \textbf{bicycle} $\uparrow$}\end{sideways}  
& \begin{sideways}{\textcolor{buscolor}{$\blacksquare$} \textbf{bus} $\uparrow$}\end{sideways}  
& \begin{sideways}{\textcolor{carcolor}{$\blacksquare$} \textbf{car} $\uparrow$}\end{sideways}  
& \begin{sideways}{\textcolor{constructcolor}{$\blacksquare$} \textbf{cons. veh.} $\uparrow$}\end{sideways}  
& \begin{sideways}{\textcolor{motorcolor}{$\blacksquare$} \textbf{motorcycle} $\uparrow$}\end{sideways}  
& \begin{sideways}{\textcolor{pedestriancolor}{$\blacksquare$} \textbf{pedestrian} $\uparrow$}\end{sideways}  
& \begin{sideways}{\textcolor{trafficcolor}{$\blacksquare$} \textbf{traffic cone} $\uparrow$}\end{sideways}  
& \begin{sideways}{\textcolor{trailercolor}{$\blacksquare$} \textbf{trailer} $\uparrow$}\end{sideways}  
& \begin{sideways}{\textcolor{truckcolor}{$\blacksquare$} \textbf{truck} $\uparrow$}\end{sideways}  
& \begin{sideways}{\textcolor{drivablecolor}{$\blacksquare$} \textbf{drive. surf.} $\uparrow$}\end{sideways}  
& \begin{sideways}{\textcolor{otherflatcolor}{$\blacksquare$} \textbf{other flat} $\uparrow$}\end{sideways}  
& \begin{sideways}{\textcolor{sidewalkcolor}{$\blacksquare$} \textbf{sidewalk} $\uparrow$}\end{sideways}  
& \begin{sideways}{\textcolor{terraincolor}{$\blacksquare$} \textbf{terrain} $\uparrow$}\end{sideways}  
& \begin{sideways}{\textcolor{manmadecolor}{$\blacksquare$} \textbf{manmade} $\uparrow$}\end{sideways}  
& \begin{sideways}{\textcolor{vegetationcolor}{$\blacksquare$} \textbf{vegetation} $\uparrow$}\end{sideways}\\ 
\midrule
% MonoScene~\cite{cao2022monoscene} & \xmark & 928 × 1600 & R101 & 6.06  & 1.75  & 7.23  & 4.26  & 4.93  & 9.38  & 5.67  & 3.98  & 3.01  & 5.90  & 4.45  & 7.17  & 14.91  & 6.32  & 7.92  & 7.43  & 1.01  & 7.65   \\
TPVFormer~\cite{huang2023tri} & \xmark & 928 × 1600 & R101 & 27.83  & 7.22  & 38.90  & 13.67  & 40.78  & 45.90  & 17.23  & 19.99  & 18.85  & 14.30  & 26.69  & 34.17  & 55.65  & 35.47  & 37.55  & 30.70  & 19.40  & 16.78  \\ 
CTF-Occ~\cite{tian2023occ3d} & \xmark & 928 × 1600 & R101 & 28.53  & 8.09  & 39.33  & 20.56  & 38.29  & 42.24  & 16.93  & 24.52  & 22.72  & 21.05  & 22.98  & 31.11  & 53.33  & 33.84  & 37.98  & 33.23  & 20.79  & 18.00    \\
OccFormer~\cite{zhang2023occformer} & \xmark & 256 × 704 & R50 & 20.40 & 6.62 & 32.57 & 13.13 & 20.37 & 37.12 & 5.04 & 14.02 & 21.01 & 16.96 & 9.34 & 20.64 & 40.89 & 27.02 & 27.43 & 18.65 & 18.78 & 16.90  \\ 
% BEVDetOcc~\cite{huang2021bevdet} & \xmark & 256 × 704 & R50 & 31.64  & 6.65  & 36.97  & 8.33  & 38.69  & 44.46  & 15.21  & 13.67  & 16.39  & 15.27  & 27.11  & 31.04  & 78.70  & 36.45  & 48.27  & 51.68  & 36.82  & 32.09  \\ 
FlashOcc (M1)~\cite{yu2023flashocc} & \xmark & 256 × 704 & R50 & 32.08  & 6.74  & 37.65  & 10.26  & 39.55  & 44.36  & 14.88  & 13.4  & 15.79  & 15.38  & 27.44  & 31.73  & 78.82  & 37.98  & 48.7  & 52.5  & 37.89  & 32.24  \\
DHD-S~\cite{wu2024deep} & \xmark & 256 × 704 & R50 & 36.50  & 10.59  & 43.21  & 23.02  & 40.61  & 47.31  & 21.68  & 23.25  & 23.85  & 23.40  & 31.75  & 34.15  & 80.16  & 41.30  & 49.95  & 54.07  & 38.73  & 33.51 \\
LightOcc-S~\cite{zhang2024lightweight} & \xmark & 256 × 704 & R50 & 37.93  & \first{11.72}  & 45.61  & \first{25.40}  & 43.10  & 48.66  & 21.38  & \first{25.58}  & 26.58  & \first{29.19}  & 33.18  & 35.09  & 79.97  & 41.81  & 50.35  & 53.88  & 39.40  & 33.97 \\
\rowcolor{gray!30}
LMPOcc-S (Ours) & \xmark & 256 × 704 & R50 & \first{40.38}  & 10.97  & \first{48.87}  & 23.66  & \first{43.31}  & \first{51.27}  & \first{23.61}  & 23.79  & \first{27.49}  & 26.28  & \first{36.26}  & \first{37.95}  & \first{81.97}  & \first{42.06}  & \first{52.09}  & \first{58.43}  & \first{52.96}  & \first{45.45} \\
\midrule

FastOcc~\cite{hou2024fastocc} & 16 & 640 × 1600 & R101 & 39.21  & 12.06 & 43.53 & 28.04 & 44.80 & 52.16 & 22.96 & 29.14 & 29.68 & 26.98 & 30.81 & 38.44 & 82.04 & 41.93 & 51.92 & 53.71 & 41.04 & 35.49  \\
PanoOcc~\cite{wang2024panoocc} & 3 & 512 × 1408 & R101 & 42.13  & 11.67 & 50.48 & 29.64  & 49.44 &  55.52  &  23.29  & 33.26 & 30.55 & 30.99 & 34.43 & 42.57 & 83.31 &  44.23 & 54.40 & 56.04 & 45.94 & 40.40  \\ 
FB-Occ~\cite{li2023fbocc} & 4 & 512 × 1408 & R101 &43.41&12.10&50.23&32.31&48.55&52.89&31.20&31.25&30.78&32.33&37.06&40.22&83.34&\first{49.27}&57.13&59.88&47.67&41.76 \\
OctreeOcc~\cite{lu2023octreeocc}& 4 & 512 × 1408 & R101 &44.02&11.96&51.70&29.93&53.52&56.77&30.83&33.17&30.65&29.99&37.76&43.87&83.17&44.52&55.45&58.86&49.52&46.33 \\
GEOcc~\cite{tan2024geocc} & 8 & 512 × 1408 & SwinB & 44.67  & 14.02  & 51.40  & 33.08  & 52.08  & 56.72  & 30.04  & 33.54  & 32.34  & 35.83  & 39.34  & 44.18  & 83.49  & 46.77  & 55.72  & 58.94  & 48.85  & 43.00 \\
COTR~\cite{ma2024cotr} & 8 & 512 × 1408 & SwinB & 46.20  & \first{14.85} & 53.25 & \first{35.19} & 50.83 & 57.25 & \first{35.36} & 34.06 & 33.54 & \first{37.14} & 38.99 & 44.97 & 84.46 & 48.73 & 57.60 & 61.08 & 51.61 & 46.72   \\ 
 \cline{2-22}

% OSP~\cite{shi2024occupancysetpoints} & 1 & 900 × 1600 & R101 &  39.41  & 11.20 & 47.25 & 27.06 & 47.57 & 53.66 & 23.21 & 29.37 & 29.68 & 28.41 & 32.39 & 39.94 & 79.35 & 41.36 & 50.31 & 53.23 & 40.52 & 35.39 \\
% BEVDetOcc-Stereo~\cite{huang2021bevdet} & 1 & 512 × 1408 & SwinB & 42.02  & 12.15 & 49.63 & 25.10 & 52.02 & 54.46 & 27.87 & 27.99 & 28.94 & 27.23 & 36.43 & 42.22 & 82.31 & 43.29 & 54.62 & 57.90 & 48.61 & 43.55  \\
FlashOcc (M3)~\cite{yu2023flashocc} & 1 & 512 × 1408 & SwinB & 43.52  & 13.42 & 51.07 & 27.68 & 51.57 & 56.22 & 27.27 & 29.98 & 29.93 & 29.80 & 37.77 & 43.52 & 83.81 & 46.55 & 56.15 & 59.56 & 50.84 & 44.67  \\
DHD-L~\cite{wu2024deep} & 1 & 512 × 1408 & SwinB & 45.53  & 14.08 & 53.12 & 32.39 & 52.44 & 57.35 & 30.83 & 35.24 & 33.01 & 33.43 & 37.90 & 45.34 & 84.61 & 47.96 & 57.39 & 60.32 & 52.27 & 46.24 \\ 

% \hline
LightOcc-L~\cite{zhang2024lightweight} & 1 & 512 × 1408 & SwinB & 46.00  & 14.50 & 52.27 & 34.45 & \first{53.79} & 57.33 & 31.80 & \first{35.83} & 33.60 & 36.09 & \first{39.89} & 46.09 & 84.23 & 48.10 & 57.14 & 60.02 & 51.70 & 45.23 \\ 
\rowcolor{gray!30}
LMPOcc-L (Ours) & 1 & 512 × 1408 & SwinB & \first{46.61}  & 13.68 & \first{53.88} & 31.65 & 53.19 & \first{58.53} & 30.68 & 34.7 & \first{34.86} & 34.6 & 39.45 & \first{47.17} & \first{85.08} & 47.85 & \first{58.11} & \first{61.56} & \first{57.36} & \first{49.97} \\ 
\bottomrule
\end{tabular}}
\label{tab:main}
\end{table*}
%%%%%%%%%%%%%%%%%%%%%%%%%%%%%%%%%%%%%%%%%%%

\subsection{Construction of 3D Open Vocabulary Maps}
\label{sec:3.4}

To build 3D open vocabulary maps for large-scale outdoor scenes, three key input modalities need to be available, namely images, their associated poses, and corresponding depth information. Dense depth estimation is crucial for accurately projecting 2D open vocabulary semantic information into the 3D space.
LMPOcc produces dense voxel occupancy grids from multi-view images, which serve as the basis for high-quality outdoor depth estimation via ray casting. 

Formally, given an image pixel \(\mathbf{u} = (u,v)^\top\), its back-projected ray direction \(\mathbf{r}_c\) in the camera coordinate system is:
\begin{equation}
\mathbf{r}_c = \mathbf{K}^{-1} {\begin{bmatrix} u,  v,  1 \end{bmatrix}}^\top\ = {\begin{bmatrix} x_c , y_c , 1 \end{bmatrix}}^\top\,
\label{eq:ray_dir}
\end{equation}
where \(\mathbf{K}\) is the camera intrinsic matrix.
Given sampled depths \(d_i = i \Delta d,\, i=1,\ldots,N_d\), 3D points along the ray in camera coordinates are computed as:
\begin{equation}
\mathbf{p}_{c,i} = d_i \mathbf{r}_c = {\begin{bmatrix} d_i x_c , d_i y_c , d_i \end{bmatrix}}^\top\,
\label{eq:points_camera}
\end{equation}
These points are transformed to ego coordinates in homogeneous form as:
%%%%%%%%%%%%%%%%%%%%%
\begin{equation}
\tilde{\mathbf{p}}_{ego,i} = \mathbf{T}_{camera \to ego} {\begin{bmatrix} \mathbf{p}_{c,i} , 1 \end{bmatrix}}^\top\ \in \mathbb{R}^4,
\label{eq:points_ego_hom}
\end{equation}
%%%%%%%%%%%%%%%%%%%%%%%%%%
where \(\mathbf{T}_{camera \to ego} \in SE(3)\) is the extrinsic transformation.
The corresponding Euclidean coordinates \(\mathbf{p}_{ego,i} \in \mathbb{R}^3\) are obtained by extracting the first three components of \(\tilde{\mathbf{p}}_{ego,i}\).
Indexing the occupancy voxel grid, voxel indices are computed by:
%%%%%%%%%%%%%%%%%%
\begin{equation}
\mathbf{v}_i = \left\lfloor \frac{\mathbf{p}_{ego,i} - \mathbf{p}_{min}}{v_{size}} \right\rfloor,
\label{eq:voxel_index}
\end{equation}
%%%%%%%%%%%%%%%%%%%
where \(\mathbf{p}_{min}\) denotes the minimum grid coordinate, and  \(v_{size}\) denotes the voxel size.
Depth at pixel \(\mathbf{u}\) is defined as the smallest sampled depth \(d_i\) along the corresponding ray for which the voxel occupancy label is not free space:
%%%%%%%%%%%%%
\begin{equation}
D(u,v) = \min \{ d_i \mid O(\mathbf{v}_i) \neq l_{\mathrm{free}} \},
\label{eq:depth_estimate}
\end{equation}
%%%%%%%%%%%%%%%%
where \(O(\cdot)\) returns voxel occupancy label and \(l_{\mathrm{free}}\) indicates free space. 
% If no occupied voxel is encountered, \(D(u,v)\) is clamped to a max depth.
In case no occupied voxel is found along the ray, the depth is set to a predefined maximum depth $D_{\max}$.

Using dense depth maps and known poses, 3D open vocabulary maps can be built with existing frameworks such as OpenGraph~\cite{deng2024opengraph}. Fig.~\ref{fig:ov_map_llm} illustrates an empirical example of how 3D open vocabulary maps enable interaction with vision-language models (VLMs) and large language models (LLMs).

\section{EXPERIMENT}
\subsection{Datasets and Metrics}
\label{sec:4.1}
We evaluate our approach on the Occ3D-nuScenes benchmark~\cite{tian2023occ3d}, which extends the widely adopted large-scale autonomous driving dataset nuScenes~\cite{caesar2020nuscenes}. This benchmark comprises 700 training scenes and 150 validation scenes, each with 40 annotated samples captured at 2Hz over 20-second sequences. The dataset spans a spatial domain of $\text{X} \in  
[-40m, 40m]$ and $\text{Y} \in [-40m, 40m]$ for horizontal dimensions, with vertical coverage from $Z \in [-1m, 5.4m]$. The occupancy annotations in this dataset are represented as axis-aligned voxels with 0.4m edge length, achieving a resolution of $200 \times 200 \times 16$ voxels. Each voxel is labeled with one of 17 semantic categories or marked as free space (non-occupied). 
In our work, $Dynamic$ represents dynamic semantic categories, which encompass $others$, $barrier$, $bicycle$, $bus$, $car$, $construction\ vehicle$, $motorcycle$, $pedestrian$, $traffic\ cone$, $trailer$ and $truck$. $Static$ denotes static semantic categories, which comprise $driveable\ surface$, $other\ flat$, $sidewalk$, $terrain$, $manmade$ and $vegetation$.
For performance evaluation, we employ the mean Intersection-Over-Union (mIoU) metric aggregated across all semantic classes.

\subsection{Implementation Details}
We employ FlashOcc~\cite{yu2023flashocc} and DHD~\cite{wu2024deep} as baseline models. During training, we initialize the models with their pretrained weights and freeze the parameters preceding the current latent feature. After integrating our LMOP module, we train for another 24 epochs while maintaining identical experimental configurations to the baseline setup. We disable BEV-space data augmentation to prevent misalignment between current features and prior features. The channel dimension of the global occupancy in LMPOcc is computed as the height of the occupancy multiplied by the number of semantic categories, specifically $16\times 18$, while other configurations regarding the global map remain consistent with those in Neural Map Prior~\cite{xiong2023neural}. All models are trained with a batch size of 4 on 6 NVIDIA A100 GPUs. The 3D open vocabulary maps are constructed using 6-view images per occupancy frame, with depth sampled at intervals $\Delta d=0.1\,m$ up to a maximum depth $D_{\max}=100\,m$.

\subsection{Main Results}
\label{sec:4.3}

%%%%%%%%%%%%%%%%%%%%%%%%%%%%%%%%%%%%%%%%%%%
\begin{table}[h!]
    \small\centering
    \caption{ The performance of occupancy prediction methods and their LMOP versions on the Occ3D-nuScenes validation set. By adding long-term memory knowledge, LMOP consistently improves these methods. }
    \resizebox{0.45\textwidth}{!}{
    \begin{tabular}{lccc}
    % \begin{tabularx}{0.8\textwidth}{>{\centering\arraybackslash}X>{\centering\arraybackslash}X>{\centering\arraybackslash}X>{\centering\arraybackslash}X>{\centering\arraybackslash}X>{\centering\arraybackslash}X}
            % \Xhline{3\arrayrulewidth}
            \toprule
    		\multirow{2}{*}{Model}&\multicolumn{3}{c}{mIoU}\\
            \cline{2-4}
    		& Dynamic & Static  & All\\
    		\midrule
    		FlashOcc-M0 &23.67 &47.11 &31.94\\
            FlashOcc-M0 + LMOP&\textbf{26.36} &\textbf{53.29} &\textbf{35.87}\\
            % \cdashline{1-5}
            % $\triangle$ mIoU&\textcolor{Maroon}{\checkmark}&\textcolor{Maroon}{+3.11}&\textcolor{Maroon}{+4.72}&\textcolor{Maroon}{+5.14}&\textcolor{Maroon}{+4.32}\\
            \rowcolor{gray!30}
            $\triangle$ mIoU &\textcolor{gray}{+2.69} &\textcolor{gray}{+6.18} &\textcolor{gray}{+3.93}\\
            
            \midrule
            
		DHD-S &29.35 &49.62 &36.5\\
		DHD-S + LMOP &\textbf{32.13} &\textbf{55.49} &\textbf{40.38}\\
            % \cdashline{1-5}
            \rowcolor{gray!30}
            $\triangle$ mIoU &\textcolor{gray}{+2.78} &\textcolor{gray}{+5.87} &\textcolor{gray}{+3.88}\\
            % \Xhline{3\arrayrulewidth}
            \bottomrule
    % \end{tabularx}
    \end{tabular}
    }
    \vspace{-0.2cm}
    \label{tab:several_basemodel}
\end{table}
%%%%%%%%%%%%%%%%%%%%%%%%%%%%%%%%%%%%%%%%%%%

Our Long-term Memory Occupancy Prior (LMOP) serves as a plug-and-play approach applicable to diverse occupancy algorithms.
 To illustrate this, we integrate LMOP into two base models: FlashOcc and DHD. We use the same hyperparameter settings as in their original designs. During training, we freeze all the modules preceding the current latent features and only train the downstream modules, especially LMOP. 
 As evidenced in Table~\ref{tab:several_basemodel}, LMOP consistently improves occupancy prediction compared to baseline models. 

We compare LMPOcc with the state-of-the-art occupancy prediction method on the Occ3D-nuScenes benchmark~\cite{tian2023occ3d}. The experiment results in Table~\ref{tab:main} show that both LMPOcc-S and LMPOcc-L outperform other approaches that have similar model settings. The baseline of LMPOcc-S and LMPOcc-L are respectively DHD-S and DHD-L.
These persuasive experiment results highlight the effectiveness of LMOP in occupancy prediction, especially on static semantic categories.

%%%%%%%%%%%%%%%%%%%%%%%%%%%%%%%%%%%%%%%%%
\begin{figure*}[h!]
    \centering
    \includegraphics[width=1\linewidth]{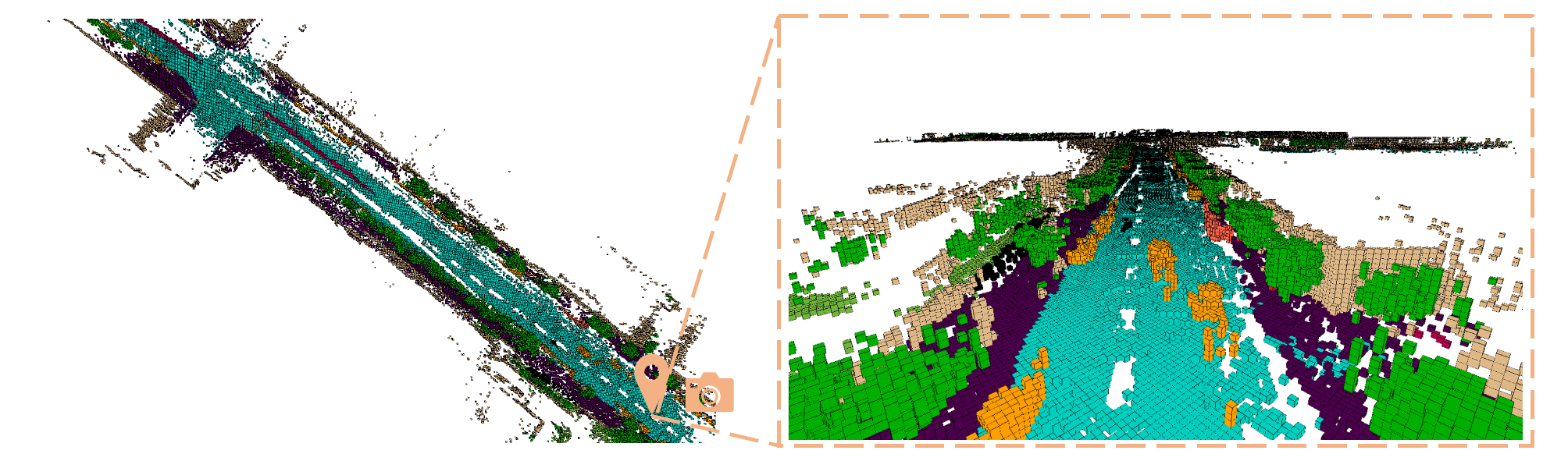}
    \vspace{-12pt}
    \caption{Visualization results of a region within our global occupancy. The left side shows the top view, and the right side shows the front view.}
    \label{fig:global_occ_vis}
\end{figure*}
%%%%%%%%%%%%%%%%%%%%%%%%%%%%%%%%%%%%%%%%%
%%%%%%%%%%%%%%%%%%%%%%%%%%%%%%%%%%%%%%%%%
\begin{figure*}[h!]
    \centering
    \includegraphics[width=\textwidth]{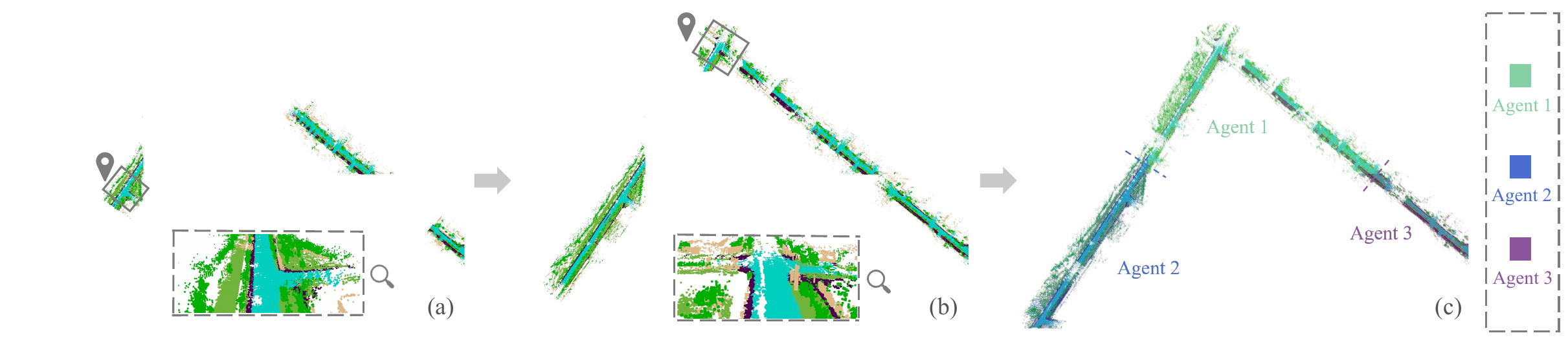}
    \vspace{-12pt}
    \caption{Visualization results of global occupancy construction via crowdsourcing methodologies. Three collaborative agents construct the global occupancy map through crowdsourcing. (a) and (b) show the intermediate stages of the occupancy construction process. (c) displays the crowdsourced mapping result. Three colors mark the areas mapped by each agent. }
    \label{fig:crowdsource}
    \vspace{-9pt}
\end{figure*}
%%%%%%%%%%%%%%%%%%%%%%%%%%%%%%%%%%%%%%%%%

\subsection{Ablation Studies}
\label{sec:4.4}
For efficient validation, we conduct ablation studies on LMPOcc-S with DHD-S as the baseline. 
The ablation studies are conducted on Current-Prior Fusion, Visibility Mask, and Dynamic Removal.

\textbf{Current-Prior Fusion.}
The performance of different fusion methods is presented in Table~\ref{tab:cpfusion}. The proposed Current-Prior Fusion module demonstrates superior performance compared to both direct concatenation and element-wise addition of current and prior features. Moreover, the joint utilization of both the Concatenation Branch and Addition Branch within the Current-Prior Fusion module yields the optimal performance. In Table~\ref{tab:cpfusion}, `C.B.' and `A.B.' represent the Concatenation Branch and Addition Branch within our Current-Prior Fusion module.

Neural Map Prior~\cite{xiong2023neural} employs cross-attention coupled with Gated Recurrent Unit (GRU) to fuse current features with prior features. As illustrated in Table~\ref{tab:atten+gru}, our Current-Prior Fusion method not only surpasses Neural Map Prior's fusion modules in performance, but also reduces computational latency, thus demonstrating significant practical advantages.

\begin{table}[h!]
\centering
\caption{Ablation study of Current-Prior Fusion. 
The term `Concat' denotes direct concatenation of current and prior features. `Add' indicates element-wise addition of the two features.
`C.B.' and `A.B.' represent the Concatenation Branch and Addition Branch within our Current-Prior Fusion module.}
\label{tab:cpfusion}
\resizebox{0.45\textwidth}{!}{%
\begin{tabular}{ccc|ccc}
\hline
\multicolumn{3}{c|}{\multirow{2}{*}{Method}}                      & \multicolumn{3}{c}{mIoU}                         \\ \cline{4-6} 
\multicolumn{3}{c|}{}                                          & Dynamic        & Static         & All            \\ \hline
\multicolumn{3}{c|}{Concat}                                    & 31.51          & 54.26          & 39.54          \\
\multicolumn{3}{c|}{Add}                                       & 31.52          & 54.41          & 39.60          \\ \hline
\multicolumn{1}{c|}{\multirow{4}{*}{CPFusion}} & C.B.   & A.B.   & \multicolumn{3}{c}{}                             \\ \cline{2-6} 
\multicolumn{1}{c|}{}                        & \cmark &        & 31.95          & 55.20          & 40.16          \\
\multicolumn{1}{c|}{}                        &        & \cmark & 31.76          & 55.09          & 40.00          \\
\multicolumn{1}{c|}{}                        & \cmark & \cmark & \textbf{32.13} & \textbf{55.49} & \textbf{40.38} \\ \hline
\end{tabular}}
\end{table}

%%%%%%%%%%%%%%%%%%%%%%%%%%%%%%%%%%%%%%
\begin{table}[h!]
    \centering
    \caption{Comparative Analysis of Fusion Modules in Neural Map Prior. Latencies are evaluated on a single A100 GPU.}
    % \scriptsize
    \resizebox{0.45\textwidth}{!}{%
    \renewcommand{\arraystretch}{1.6}
    \begin{tabular}{lcc}
        \toprule
        Method & mIoU$\uparrow$ & Latency$\downarrow$\\ 
        \hline
        Cross Attention + GRU~\cite{xiong2023neural}& 39.80 & 11.6 ms\\
        CPFusion (Ours) & \textbf{40.38} & \textbf{7.1 ms}\\
        \bottomrule
    \end{tabular}}
    \label{tab:atten+gru}
\end{table}
%%%%%%%%%%%%%%%%%%%%%%%%%%%%%%%%%%%%%%

%%%%%%%%%%%%%%%%%%%%%%%%%%%%%%%%%%%%%%%%%%%
\begin{table}[h!]
    \centering
    \caption{ Ablation study of Camera Visibility Mask. }
    \resizebox{0.45\textwidth}{!}{
    \begin{tabular}{lccc}
    % \begin{tabularx}{0.8\textwidth}{>{\centering\arraybackslash}X>{\centering\arraybackslash}X>{\centering\arraybackslash}X>{\centering\arraybackslash}X>{\centering\arraybackslash}X>{\centering\arraybackslash}X}
            % \Xhline{3\arrayrulewidth}
            \toprule
    		\multirow{2}{*}{Method}&\multicolumn{3}{c}{mIoU}\\
            \cline{2-4}
    		& Dynamic & Static  & All\\
            
            \midrule
            Baseline &29.35 &49.62 &36.5\\
		Baseline + LMOP (w/o Mask) &28.87     &49.43     &36.13    \\
		Baseline + LMOP (w/ Mask) &\textbf{32.13} &\textbf{55.49} &\textbf{40.38}\\
            \bottomrule
    % \end{tabularx}
    \end{tabular}
    }
    \vspace{-0.2cm}
    \label{tab:camera_mask}
\end{table}

%%%%%%%%%%%%%%%%%%%%%%%%%%%%%%%%%%%%%%%%%%%

%%%%%%%%%%%%%%%%%%%%%%%%%%%%%%%%%%%%%%%%%%%
\begin{table}[h!]
    \centering
    \caption{ Analysis of Dynamic Targets in LMOP. }
    \resizebox{0.45\textwidth}{!}{
    \begin{tabular}{lccc}
    % \begin{tabularx}{0.8\textwidth}{>{\centering\arraybackslash}X>{\centering\arraybackslash}X>{\centering\arraybackslash}X>{\centering\arraybackslash}X>{\centering\arraybackslash}X>{\centering\arraybackslash}X}
            % \Xhline{3\arrayrulewidth}
            \toprule
    		\multirow{2}{*}{Method}&\multicolumn{3}{c}{mIoU}\\
            \cline{2-4}
    		& Dynamic & Static  & All\\
            
            \midrule
            Removing Dynamic v1 &30.45     &55.28     &39.21\\
		Removing Dynamic v2 &32.05     &55.29     &40.25   \\
		Retaining Dynamic &\textbf{32.13} &\textbf{55.49} &\textbf{40.38}\\
            \bottomrule
    % \end{tabularx}
    \end{tabular}
    }
    \vspace{-0.2cm}
    \label{tab:remove_dynamic}
\end{table}
%%%%%%%%%%%%%%%%%%%%%%%%%%%%%%%%%%%%%%%%%%%

\textbf{Visibility Mask.}
We analyze the impact of the visibility mask within our framework. As demonstrated in Table~\ref{tab:camera_mask}, LMPOcc slightly underperforms the baseline without the visibility mask, whereas its performance significantly surpasses the baseline when the visibility mask is applied. This is because regions outside the visibility mask contain substantial noise, thus storing information exclusively within the visible regions ensures the validity of the prior.

%%%%%%%%%%%%%%%%%%%%%%%%%%%%%%%%%%%%%%%%%%
\begin{figure}[t]
    \centering
    \Huge
    \includegraphics[width=\columnwidth]{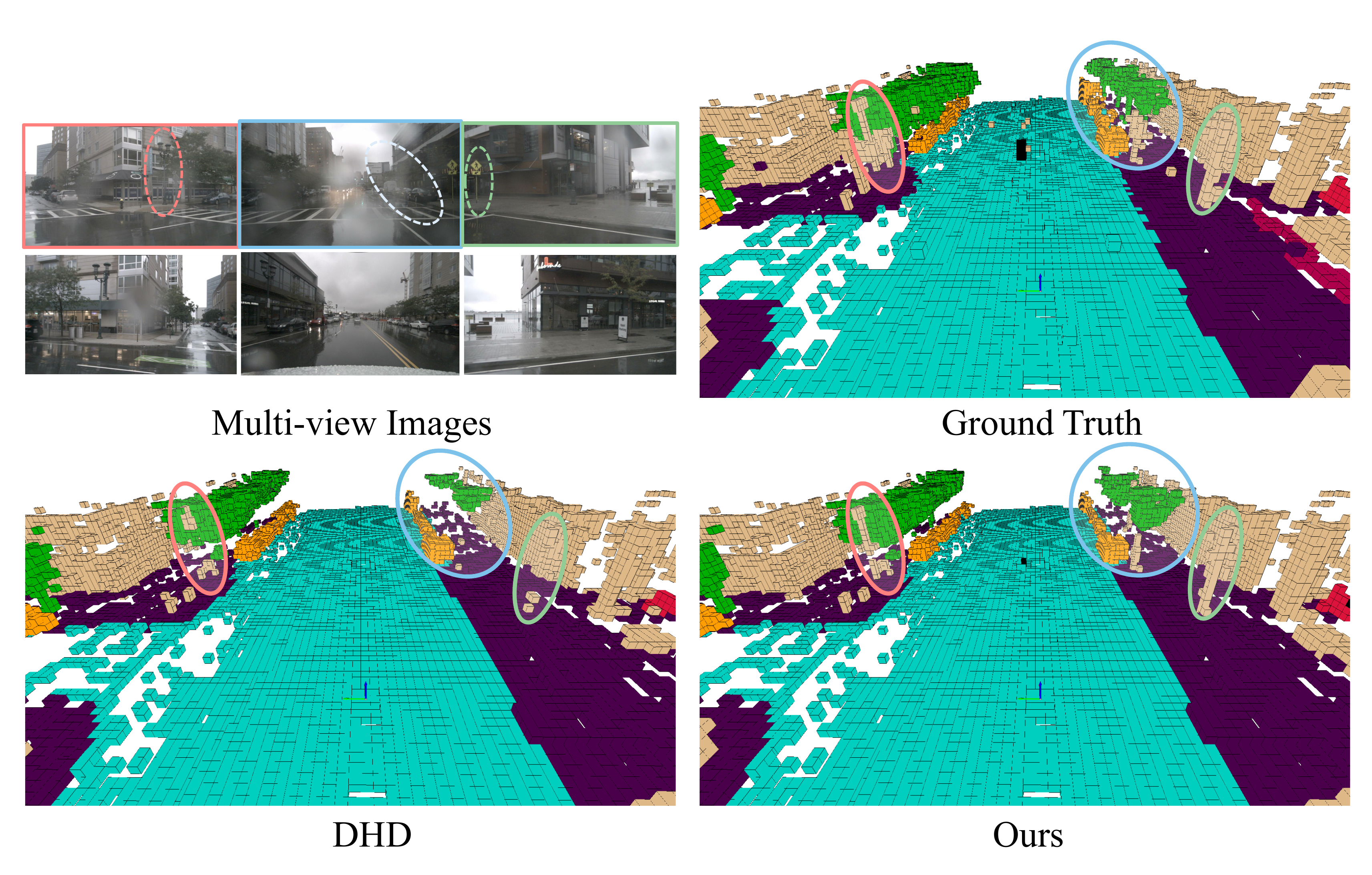}
    \vspace{-30pt}
    \caption{Visualization results of LMPOcc and DHD. It is a low-visibility rainy scene. LMPOcc leverages long-term memory priors to detect objects not visible in current sensory observations, demonstrating significant improvement over the baseline. The color-semantic category correspondence is detailed in Table~\ref{tab:main}.}
    \label{fig:visualization}
    \vspace{-9pt}
\end{figure}
%%%%%%%%%%%%%%%%%%%%%%%%%%%%%%%%%%%%%%%%%%

%%%%%%%%%%%%%%%%%%%%%%%%%%%%%%%%%%%%%%%%%%
\begin{figure}[t]
    \centering
    \Huge
    \includegraphics[width=\columnwidth]{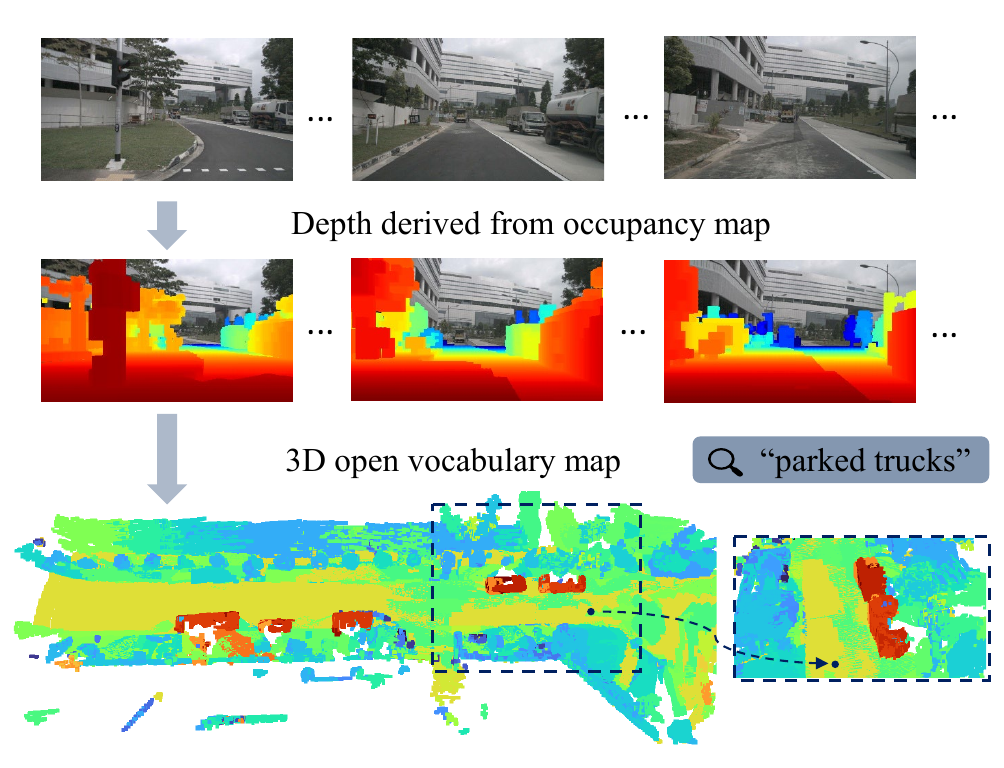}
    \vspace{-28pt}
    % \caption{Visualization results of 3D open vocabulary map. It derives depth information from the occupancy map and builds upon the OpenGraph method to construct the outdoor open-vocabulary map. The bottom row demonstrates scene-level occupancy grounding.}
     \caption{Visualization results of the 3D open vocabulary map. Depth information is derived from the occupancy map, and the outdoor open-vocabulary map is constructed based on the OpenGraph method. The bottom row shows scene-level occupancy grounding for specific queries (e.g., “parked trucks”).}
    \label{fig:query_ov_map}
    \vspace{-9pt}
\end{figure}
%%%%%%%%%%%%%%%%%%%%%%%%%%%%%%%%%%%%%%%%%%

\textbf{Discussion of Dynamic Targets.}
Given the temporal inconsistencies of dynamic targets at the same geographic location across distinct observation periods, we conduct analysis on the handling of dynamic targets within the global prior in our framework. We propose two distinct methodologies for eliminating dynamic elements within the prior framework. After applying the argmax operation to the occupancy logits, the semantic label for each voxel can be obtained. We create free masks and dynamic masks to, respectively, mask out the free components and dynamic components from the occupancy logits. 
The dynamic components comprise dynamic semantic categories, as detailed in Sec. \ref{sec:4.1}.
The first method sets the dynamic components in the occupancy logits as zero, denoted as $Removing\ Dynamic\ v1$ in Table~\ref{tab:remove_dynamic}. The second method randomly selects the free components in the occupancy logits and replaces the dynamic components with them, denoted as $Removing\ Dynamic\ v2$ in Table~\ref{tab:remove_dynamic}. 

The results indicate that removing dynamic elements does not yield performance improvements, as detailed in Table~\ref{tab:remove_dynamic}. A plausible explanation lies in our Current-Prior Fusion module's capability to adaptively process dynamic components within prior features. Another critical factor is that dynamic objects typically exhibit spatial-temporal distribution patterns in specific regions, which serves as effective prior knowledge to enhance dynamic object detection. This mirrors how the human brain leverages spatial priors in perceptual processing to identify moving entities.

\subsection{Visualization}

Our global occupancy is shown in Fig.~\ref{fig:global_occ_vis}. While enhancing local perception capabilities, LMPOcc can construct global occupancy for large-scale scenes. As shown in Fig.~\ref{fig:crowdsource}, three collaborative agents construct the global occupancy map through crowdsourcing. The visualization results of LMPOcc-S are presented in Fig.~\ref{fig:visualization}. It shows a rainy, low-visibility scene where LMPOcc leverages long-term memory priors to detect objects missing in current views, significantly outperforming the baseline.

\subsection{Construction of 3D Open Vocabulary Maps}

In Fig.~\ref{fig:query_ov_map}, we showcase the results of a 3D open-vocabulary map constructed with OpenGraph~\cite{deng2024opengraph}, utilizing depth generated from the occupancy map for a sequence of multi-view images. The first row shows the image sequence used for building the 3D map, the second row illustrates the depth extracted from the occupancy map for each image via ray casting, and the third row displays a scene-level 3D open-vocabulary map along with the grounding of "parked trucks" on the map.

\section{CONCLUSION}

We propose LMPOcc, a novel 3D occupancy prediction framework that leverages long-term memory priors from historical traversals to enhance local perception and build unified global occupancy maps. By introducing a model-agnostic prior format, LMPOcc ensures compatibility across different prediction baselines, while a lightweight Current-Prior Fusion module adaptively integrates prior and current features. Validated on the Occ3D-nuScenes benchmark, LMPOcc achieves state-of-the-art local occupancy prediction, especially for static semantic classes. Additionally, it supports large-scale global occupancy construction via multi-vehicle crowdsourcing and facilitates 3D open-vocabulary map building through occupancy-derived dense depth.
Future research could conduct in-depth investigations into dynamic object processing within prior information and develop learnable optimization frameworks for global occupancy.

\bibliographystyle{IEEEtran}
\bibliography{ref}

@inproceedings{cao2022monoscene,
  title={Monoscene: Monocular 3d semantic scene completion},
  author={Cao, Anh-Quan and De Charette, Raoul},
  booktitle={Proceedings of the IEEE/CVF Conference on Computer Vision and Pattern Recognition},
  pages={3991--4001},
  year={2022}
}

@inproceedings{ye2024cvt,
  title={Cvt-occ: Cost volume temporal fusion for 3d occupancy prediction},
  author={Ye, Zhangchen and Jiang, Tao and Xu, Chenfeng and Li, Yiming and Zhao, Hang},
  booktitle={European Conference on Computer Vision},
  pages={381--397},
  year={2024},
  organization={Springer}
}

@article{park2022time,
  title={Time will tell: New outlooks and a baseline for temporal multi-view 3d object detection},
  author={Park, Jinhyung and Xu, Chenfeng and Yang, Shijia and Keutzer, Kurt and Kitani, Kris and Tomizuka, Masayoshi and Zhan, Wei},
  journal={arXiv preprint arXiv:2210.02443},
  year={2022}
}

@article{huang2022bevdet4d,
  title={Bevdet4d: Exploit temporal cues in multi-camera 3d object detection},
  author={Huang, Junjie and Huang, Guan},
  journal={arXiv preprint arXiv:2203.17054},
  year={2022}
}

@inproceedings{li2023voxformer,
  title={Voxformer: Sparse voxel transformer for camera-based 3d semantic scene completion},
  author={Li, Yiming and Yu, Zhiding and Choy, Christopher and Xiao, Chaowei and Alvarez, Jose M and Fidler, Sanja and Feng, Chen and Anandkumar, Anima},
  booktitle={Proceedings of the IEEE/CVF conference on computer vision and pattern recognition},
  pages={9087--9098},
  year={2023}
}

@article{li2024bevformer,
  title={Bevformer: learning bird's-eye-view representation from lidar-camera via spatiotemporal transformers},
  author={Li, Zhiqi and Wang, Wenhai and Li, Hongyang and Xie, Enze and Sima, Chonghao and Lu, Tong and Yu, Qiao and Dai, Jifeng},
  journal={IEEE Transactions on Pattern Analysis and Machine Intelligence},
  year={2024},
  publisher={IEEE}
}

@article{zhang2023occnerf,
  title={Occnerf: Self-supervised multi-camera occupancy prediction with neural radiance fields},
  author={Zhang, Chubin and Yan, Juncheng and Wei, Yi and Li, Jiaxin and Liu, Li and Tang, Yansong and Duan, Yueqi and Lu, Jiwen},
  journal={CoRR},
  year={2023}
}

@inproceedings{huang2024selfocc,
  title={Selfocc: Self-supervised vision-based 3d occupancy prediction},
  author={Huang, Yuanhui and Zheng, Wenzhao and Zhang, Borui and Zhou, Jie and Lu, Jiwen},
  booktitle={Proceedings of the IEEE/CVF Conference on Computer Vision and Pattern Recognition},
  pages={19946--19956},
  year={2024}
}

@inproceedings{wei2023surroundocc,
  title={Surroundocc: Multi-camera 3d occupancy prediction for autonomous driving},
  author={Wei, Yi and Zhao, Linqing and Zheng, Wenzhao and Zhu, Zheng and Zhou, Jie and Lu, Jiwen},
  booktitle={Proceedings of the IEEE/CVF International Conference on Computer Vision},
  pages={21729--21740},
  year={2023}
}

@inproceedings{huang2023tri,
  title={Tri-perspective view for vision-based 3d semantic occupancy prediction},
  author={Huang, Yuanhui and Zheng, Wenzhao and Zhang, Yunpeng and Zhou, Jie and Lu, Jiwen},
  booktitle={Proceedings of the IEEE/CVF conference on computer vision and pattern recognition},
  pages={9223--9232},
  year={2023}
}

@inproceedings{philion2020lift,
  title={Lift, splat, shoot: Encoding images from arbitrary camera rigs by implicitly unprojecting to 3d},
  author={Philion, Jonah and Fidler, Sanja},
  booktitle={Computer Vision--ECCV 2020: 16th European Conference, Glasgow, UK, August 23--28, 2020, Proceedings, Part XIV 16},
  pages={194--210},
  year={2020},
  organization={Springer}
}

@inproceedings{zhang2023occformer,
  title={Occformer: Dual-path transformer for vision-based 3d semantic occupancy prediction},
  author={Zhang, Yunpeng and Zhu, Zheng and Du, Dalong},
  booktitle={Proceedings of the IEEE/CVF International Conference on Computer Vision},
  pages={9433--9443},
  year={2023}
}

@article{huang2021bevdet,
  title={Bevdet: High-performance multi-camera 3d object detection in bird-eye-view},
  author={Huang, Junjie and Huang, Guan and Zhu, Zheng and Ye, Yun and Du, Dalong},
  journal={arXiv preprint arXiv:2112.11790},
  year={2021}
}

@article{yu2023flashocc,
  title={Flashocc: Fast and memory-efficient occupancy prediction via channel-to-height plugin},
  author={Yu, Zichen and Shu, Changyong and Deng, Jiajun and Lu, Kangjie and Liu, Zongdai and Yu, Jiangyong and Yang, Dawei and Li, Hui and Chen, Yan},
  journal={arXiv preprint arXiv:2311.12058},
  year={2023}
}

@article{wu2024deep,
  title={Deep height decoupling for precise vision-based 3d occupancy prediction},
  author={Wu, Yuan and Yan, Zhiqiang and Wang, Zhengxue and Li, Xiang and Hui, Le and Yang, Jian},
  journal={arXiv preprint arXiv:2409.07972},
  year={2024}
}

@article{zhang2024lightweight,
  title={Lightweight Spatial Embedding for Vision-based 3D Occupancy Prediction},
  author={Zhang, Jinqing and Zhang, Yanan and Liu, Qingjie and Wang, Yunhong},
  journal={arXiv preprint arXiv:2412.05976},
  year={2024}
}

@article{hou2024fastocc,
  title={FastOcc: Accelerating 3D Occupancy Prediction by Fusing the 2D Bird's-Eye View and Perspective View},
  author={Hou, Jiawei and Li, Xiaoyan and Guan, Wenhao and Zhang, Gang and Feng, Di and Du, Yuheng and Xue, Xiangyang and Pu, Jian},
  journal={arXiv preprint arXiv:2403.02710},
  year={2024}
}

@inproceedings{wang2024panoocc,
  title={Panoocc: Unified occupancy representation for camera-based 3d panoptic segmentation},
  author={Wang, Yuqi and Chen, Yuntao and Liao, Xingyu and Fan, Lue and Zhang, Zhaoxiang},
  booktitle={Proceedings of the IEEE/CVF conference on computer vision and pattern recognition},
  pages={17158--17168},
  year={2024}
}

@article{li2023fbocc,
  title={Fb-occ: 3d occupancy prediction based on forward-backward view transformation},
  author={Li, Zhiqi and Yu, Zhiding and Austin, David and Fang, Mingsheng and Lan, Shiyi and Kautz, Jan and Alvarez, Jose M},
  journal={arXiv preprint arXiv:2307.01492},
  year={2023}
}

@article{lu2023octreeocc,
  title={OctreeOcc: Efficient and multi-granularity occupancy prediction using octree queries},
  author={Lu, Yuhang and Zhu, Xinge and Wang, Tai and Ma, Yuexin},
  journal={arXiv preprint arXiv:2312.03774},
  year={2023}
}

@article{tan2024geocc,
  title={GEOcc: Geometrically Enhanced 3D Occupancy Network with Implicit-Explicit Depth Fusion and Contextual Self-Supervision},
  author={Tan, Xin and Wu, Wenbin and Zhang, Zhiwei and Fan, Chaojie and Peng, Yong and Zhang, Zhizhong and Xie, Yuan and Ma, Lizhuang},
  journal={arXiv preprint arXiv:2405.10591},
  year={2024}
}

@inproceedings{ma2024cotr,
  title={Cotr: Compact occupancy transformer for vision-based 3d occupancy prediction},
  author={Ma, Qihang and Tan, Xin and Qu, Yanyun and Ma, Lizhuang and Zhang, Zhizhong and Xie, Yuan},
  booktitle={Proceedings of the IEEE/CVF Conference on Computer Vision and Pattern Recognition},
  pages={19936--19945},
  year={2024}
}

@inproceedings{xiong2023neural,
  title={Neural map prior for autonomous driving},
  author={Xiong, Xuan and Liu, Yicheng and Yuan, Tianyuan and Wang, Yue and Wang, Yilun and Zhao, Hang},
  booktitle={Proceedings of the IEEE/CVF Conference on Computer Vision and Pattern Recognition},
  pages={17535--17544},
  year={2023}
}

@article{tian2023occ3d,
  title={Occ3d: A large-scale 3d occupancy prediction benchmark for autonomous driving},
  author={Tian, Xiaoyu and Jiang, Tao and Yun, Longfei and Mao, Yucheng and Yang, Huitong and Wang, Yue and Wang, Yilun and Zhao, Hang},
  journal={Advances in Neural Information Processing Systems},
  volume={36},
  pages={64318--64330},
  year={2023}
}

@inproceedings{caesar2020nuscenes,
  title={nuscenes: A multimodal dataset for autonomous driving},
  author={Caesar, Holger and Bankiti, Varun and Lang, Alex H and Vora, Sourabh and Liong, Venice Erin and Xu, Qiang and Krishnan, Anush and Pan, Yu and Baldan, Giancarlo and Beijbom, Oscar},
  booktitle={Proceedings of the IEEE/CVF conference on computer vision and pattern recognition},
  pages={11621--11631},
  year={2020}
}

@article{deng2024opengraph,
  title={Opengraph: Open-vocabulary hierarchical 3d graph representation in large-scale outdoor environments},
  author={Deng, Yinan and Wang, Jiahui and Zhao, Jingyu and Tian, Xinyu and Chen, Guangyan and Yang, Yi and Yue, Yufeng},
  journal={IEEE Robotics and Automation Letters},
  year={2024},
  publisher={IEEE}
}

@article{leng2025occupancy,
  title={Occupancy Learning with Spatiotemporal Memory},
  author={Leng, Ziyang and Yang, Jiawei and Yi, Wenlong and Zhou, Bolei},
  journal={arXiv preprint arXiv:2508.04705},
  year={2025}
}

@article{pavkovic2025gaussianfusionocc,
  title={GaussianFusionOcc: A Seamless Sensor Fusion Approach for 3D Occupancy Prediction Using 3D Gaussians},
  author={Pavkovi{\'c}, Tomislav and Mahani, Mohammad-Ali Nikouei and Niedermayer, Johannes and Betz, Johannes},
  journal={arXiv preprint arXiv:2507.18522},
  year={2025}
}

\end{document}